\documentclass[letterpaper]{article} 
\usepackage{aaai25}  
\usepackage{times}  
\usepackage{helvet}  
\usepackage{courier}  
\usepackage[hyphens]{url}  
\usepackage{graphicx} 
\usepackage{natbib}  
\usepackage{caption} 
\usepackage{algorithm}
\usepackage{algorithmic}
\usepackage[table]{xcolor}
\usepackage{multirow}
\usepackage{amssymb}
\usepackage{amsmath}
\usepackage{newfloat}
\usepackage{listings}
\DeclareCaptionStyle{ruled}{labelfont=normalfont,labelsep=colon,strut=off} 
\floatstyle{ruled}
\newfloat{listing}{tb}{lst}{}
\floatname{listing}{Listing}
\title{Radiology Report Generation via Multi-objective Preference Optimization}
\author {
     Ting Xiao\textsuperscript{\rm 1},
     Lei Shi\textsuperscript{\rm 1},
     Peng Liu\textsuperscript{\rm 2},
     Zhe Wang\textsuperscript{\rm 1}\thanks{Corresponding author.},
     Chenjia Bai$^{3*}$ 
}

\affiliations {
    \textsuperscript{\rm 1}East China University of Science and Technology\\
    \textsuperscript{\rm 2}Harbin Institute of Technology\\
    \textsuperscript{\rm 3}Institute of Artificial Intelligence (TeleAI), China Telecom\\
   xiaoting@ecust.edu.cn, y80230058@mail.ecust.edu.cn, egami@126.com,
   wangzhe@ecust.edu.cn, baicj@chinatelecom.cn
}

\begin{document}

\maketitle

\begin{abstract}
Automatic Radiology Report Generation (RRG) is an important topic for alleviating the substantial workload of radiologists. Existing RRG approaches rely on supervised regression based on different architectures or additional knowledge injection, while the generated report may not align optimally with radiologists’ preferences. Especially, since the preferences of radiologists are inherently heterogeneous and multi-dimensional, e.g., some may prioritize report fluency, while others emphasize clinical accuracy. To address this problem, we propose a new RRG method via Multi-objective Preference Optimization (MPO) to align the pre-trained RRG model with multiple human preferences, which can be formulated by multi-dimensional reward functions and optimized by multi-objective reinforcement learning (RL). Specifically, we use a preference vector to represent the weight of preferences and use it as a condition for the RRG model. Then, a linearly weighed reward is obtained via a dot product between the preference vector and multi-dimensional reward. Next, the RRG model is optimized to align with the preference vector by optimizing such a reward via RL. In the training stage, we randomly sample diverse preference vectors from the preference space and align the model by optimizing the weighted multi-objective rewards, which leads to an optimal policy on the entire preference space. When inference, our model can generate reports aligned with specific preferences without further fine-tuning. Extensive experiments on two public datasets show the proposed method can generate reports that cater to different preferences in a single model and achieve state-of-the-art performance.
\end{abstract}

%

\section{Introduction}
Radiology reports provide an important basis for physicians to make diagnoses and are usually written by experienced radiologists based on the syndromes observed in medical images. However, manually writing reports is labor-intensive, time-consuming, and error-prone due to the large amount of daily medical reports and variations in experience among radiologists. Therefore, automatic radiology report generation (RRG) relying on deep learning techniques as an alternative has emerged as an attractive research topic in both artificial intelligence and medicine. 

To generate high-quality reports, a series of approaches \cite{ R2Gen, liu2024multi} have been proposed. They usually adopt the classic encoder-decoder structure and generate better text descriptions by promoting the model’s structure, such as introducing a cross-modal memory module \cite{R2GenCMN, MAN}, a multi-modal alignment module \cite{yang2023radiology}, or new attention mechanisms \cite{CMCA}. Another line of approaches improves report generation by injecting additional knowledge, such as knowledge graphs \cite{yang2022knowledge, kale2023knowledge, KIUT} or disease tags \cite{aligntransformer, wang2022medical, jin2024promptmrg}, or retrieved reports from other corpora \cite{liu2024bootstrapping}. 

The above RRG methods rely on supervised regression based on different architectures or additional knowledge injection. However, the generated reports may not align optimally with radiologists’ preferences. Some approaches \cite{qin2022reinforced, delbrouck2022improving} integrate reinforcement learning (RL) into report generation to maximize rewards that assess report quality. These RL-based approaches generally use a single evaluation metric or a simple weighted sum of multiple metrics to formulate the reward function. This function typically represents a weighted aggregation of multiple objectives, tailored to a specific preference. Nonetheless, preferences are inherently heterogeneous and multi-dimensional. For instance, some radiologists may prioritize the fluency of the text, while others may focus on clinical accuracy. Therefore, RRG models must cater to the diverse preferences of different radiologists, a challenge that is difficult to achieve with a single model and remains unexplored in previous works. 

In this paper, we study this unexplored topic for the first time and propose a new RRG method to align the pre-trained RRG model with multiple human preferences via Multi-objective Preference Optimization (MPO). The key challenge is how to utilize a low-dimensional preference vector to control the model's behavior. To address this challenge, MPO formalizes the preferences as a multi-dimensional vector and uses it as a condition for the RRG model, which can be formulated by multi-dimensional reward functions and optimized by multi-objective reinforcement learning. Specifically, our MPO introduces two new modules: the preference vector fusion (PVF) network and the multi-objective optimization (MOO) module. The PVF network is situated between the encoder and decoder of the standard Transformer. It employs a multi-head attention mechanism and residual connections to fuse the preference vector with the encoded image features, enabling conditional generation. The MOO module uses the preference vector to represent the weight of the preference and linearly weights the multi-dimensional reward with the preference vector. Then, RL is used to optimize the weighted reward function, promoting the RRG model to align with the preference. During training, we randomly sample diverse preference vectors from the preference space and align the RRG model by optimizing the weighted multi-objective rewards functions, which leads to an optimal policy on the entire preference space, ensuring the optimal RRG model performs differently under various preferences. During inference, our model can generate reports aligned with specific preferences without further fine-tuning. Our main contribution is summarized as follows:
\begin{itemize}
    \item We propose a new RRG method that aligns with human preferences and formulate it as a multi-objective optimization reinforcement learning problem. To the best of our knowledge, this is the first study to align an RRG model with human preferences.
    \item We propose a preference vector fusion module that fuses the encoded image features with the preference vector using an attention mechanism and residual connections, ensuring that the preference vector can condition the report generation.
    \item Extensive experiments on two public datasets demonstrate that our proposed method can cater to different preferences within a single model without further fine-tuning and achieve state-of-the-art performance.
\end{itemize}

\section{Related Works}\label{section 2}

\subsection{Radiology Report Generation }
Recent advancements in RRG can be classified into two main strategies: improving model architectures and injecting additional knowledge.

\subsubsection{Improving Model Architectures} 
Early RRG studies are implemented on CNN-RNN or CNN-LSTM architectures \cite{wang2020unifying, najdenkoska2021variational}. Recent works have focused on improving the network based on the CNN-Transformer architecture. For instance, both CMN \cite{R2GenCMN} and MAN \cite{MAN} introduce a cross-modal memory network featuring a shared memory that records the alignment information between images and text. XPRONET \cite{XPRONET} proposes a cross-modal prototype-driven network to enhance cross-modal pattern learning. METrans \cite{Metransfor} introduces multiple learnable ``expert'' tokens into both the Transformer encoder and decoder, enhancing image representation and report generation. To capture subtle differences in radiology images, various attention mechanisms have been developed to enhance feature robustness in report generation, such as the co-attention mechanism \cite{jing2018automatic}. Additionally, Ali-Transformer \cite{aligntransformer} introduces cross-view attention, using single-view and multi-view branches to combine features from frontal and lateral images with squeeze-excitation attention to aid report generation. \citet{CMCA} introduce a cross-modal contrastive attention (CMCA) model that uses contrastive features from similar historical cases to guide report generation. \citet{wang2022medical} propose a memory-augmented sparse attention block that integrates bilinear pooling with self-attention to capture higher-order interactions and generate long reports.

\subsubsection{Injecting Additional Knowledge } 
This type of work explores the injection of additional knowledge, such as disease labels \cite{tanida2023interactive, jin2024promptmrg}, retrieved reports \cite{liu2024bootstrapping}, and knowledge graphs \cite{kale2023knowledge, KIUT}, to assist in report generation. PPKED \cite{liu2021exploring} combines abnormal findings, knowledge graphs, and retrieved reports to mimic the working patterns of radiologists. 
KiUT \cite{KIUT} enhances results by integrating visual and contextual knowledge with external clinical insights through an injected knowledge distiller. RGRG \cite{tanida2023interactive} uses an object detector as region guidance for report generation, allowing the decoder to directly utilize disease information. \citet{jin2024promptmrg} generates high-quality reports by converting diagnostic results from a disease classification branch and combining them with retrieved reports.

\begin{figure*}[htp]
\centering
\includegraphics[width=0.8\textwidth]{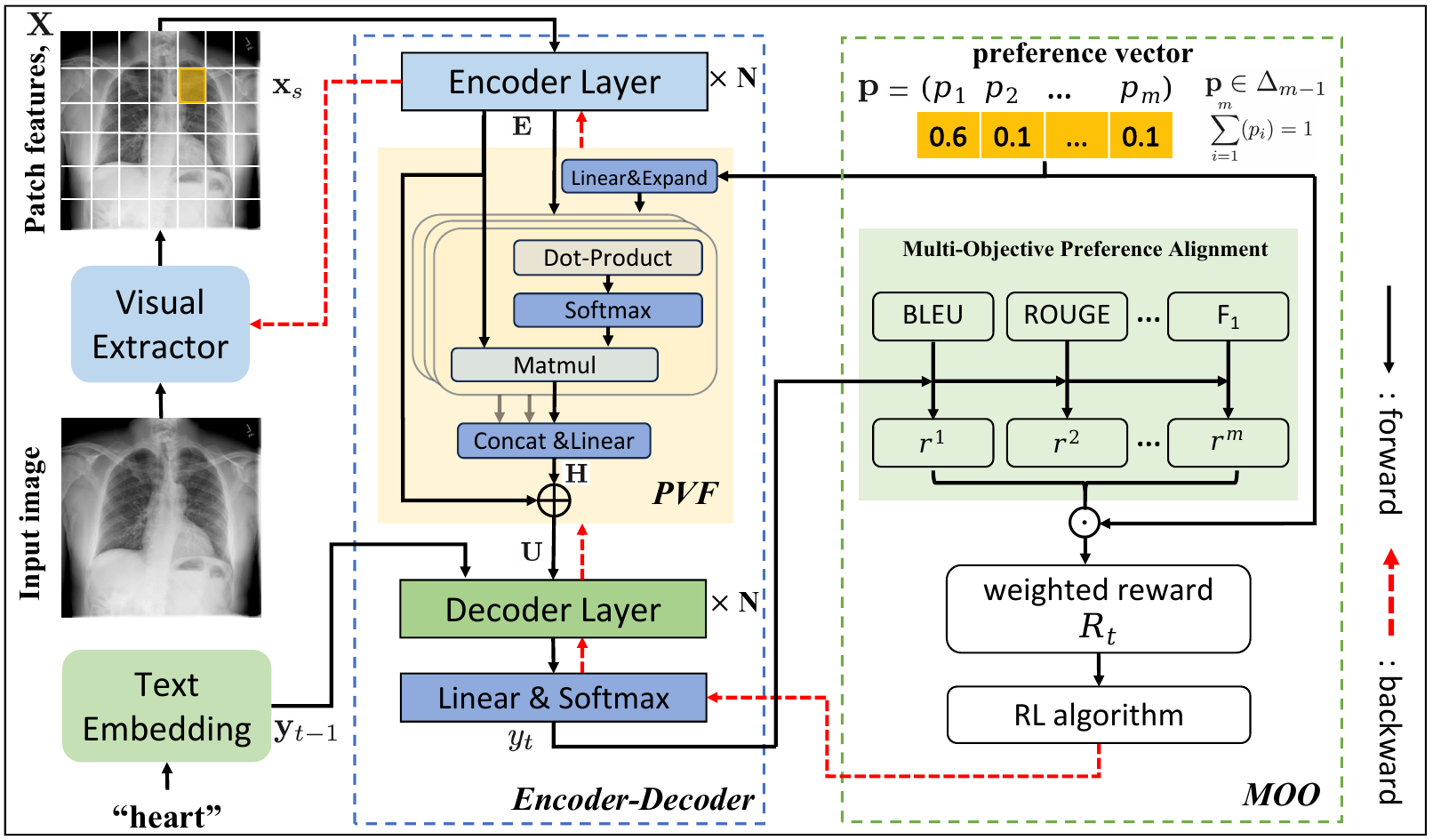}
\caption{The architecture of our MPO, where the blue dashed box represents the encoder-decoder with the PVF network, and the green dashed box represents the MOO module. The red dashed arrows represent the back-propagation of gradients.}
\label{fig1}
\end{figure*}

\subsection{RRG via Reinforcement Learning}
The aforementioned RRG methods are primarily single-stage and rely on supervised regression. Recently, reinforcement learning (RL) has been integrated into RRG tasks, further enhancing report-generation capabilities through appropriate supervision from carefully designed rewards. These rewards are typically calculated based on natural language generation (NLG) metrics \cite{wang2021self, wang2022medical} or semantic relevance metrics \cite{delbrouck2022improving}. For example, CMN+RL \cite{qin2022reinforced} leverages signal from NLG metrics, such as BLEU, as a reward to guide the cross-modal mappings between image and text features. \citet{li2018hybrid} design a hierarchical decision-making policy guided by sentence-level and word-level rewards to determine whether to retrieve a template sentence or generate a new one. \citet{miura2021improving} design two new rewards, namely, the exact entity matching reward and the entity matching reward, based on entity coverage and relationships. These rewards are combined with NLG metric rewards to improve the factual completeness and consistency of generated reports. Similarly, \citet{delbrouck2022improving} employ entities and relationships extracted by PubMedBERT \cite{PubMedBERT} to compute three rewards to enhance the factual completeness and correctness of report generation. \citet{wu2023token} introduce an imbalanced evaluation reward that divides performance differences into four groups based on token frequency, balancing performance to ensure that tokens from these groups are given equal weight. 

These RL-based report generation methods typically employ either a single evaluation metric or a straightforward weighted sum of multiple metrics as the reward function. Such a reward function represents a weighted aggregation of multiple objectives under a specific preference vector, which is essentially a single-objective optimization and can only produce a policy applicable to a particular preference. Consequently, the model must be retrained to adapt to different preferences. Recent studies \cite{yu2024regularized,yuan2024preference,shan2024forward} incorporate preference information into the generation process to guide preference alignment.
Our approach is orthogonal to the above methods, and we are the first to align preferences with the RRG model via multi-objective preference optimization.

%
\pdfinfo{
/TemplateVersion (2025.1)
}

\section{Methods}
\subsection{Overview and Generation Pipeline}
We aim to train a model that can generate reports aligned with human preferences. As shown in Figure \ref{fig1}, our MPO builds upon the mainstream encoder-decoder framework and comprises two novel modules: the encoder-decoder with our preference vector fusion (PVF) network, and the multi-objective optimization (MOO) module. In our framework, RRG is also formulated as a sequence-to-sequence task, where, given a radiology image $I$, the Encoder Layers encode the visual features extracted by the visual extractor and the Decoder Layers generate report $Y = \left[y_1, y_2, \ldots, y_t, \ldots, y_T\right]$ conditioned on both visual features and preference vector $\mathbf{p}$. Here, $ y_t \in \mathbb{V}$ is the generated tokens at step $t$, $T$ is the length of the report, and $\mathbb{V}$ represents the vocabulary space. The entire generation process can be formulated as follows:
\begin{equation}
   p(Y \mid I,\mathbf{p}) = \prod \limits_{t=1}^T p(y_t \mid y_1, \ldots, y_{t-1}, I, \mathbf{p}) .
\end{equation}

Firstly, we extract visual features $\mathbf{X}$ from radiology image $I$ via a pre-trained CNN, specifically ResNet 101 \cite{he2016deep}. Then, decompose the image into equally sized patches, with the patch features being extracted from the last convolutional layer of the CNN. These patch features are then expanded into a sequence by concatenating them row-by-row, which is formulated as follows:
\begin{equation}
    \mathbf{X} = \left[ \mathbf{x}_1, \mathbf{x}_2, \ldots, \mathbf{x}_s, \ldots, \mathbf{x}_S \right] = f_v(I),
\end{equation}
where $f_v$ is the visual extractor, $\mathbf{x}_s \in \mathbb{R}^d$ is a patch feature, $d$ is the feature dimension, and $S$ represents the patch number.

Then, the extracted visual features $\mathbf{X} $ and text embedding $\left[ \mathbf{y}_1, \mathbf{y}_2, \ldots, \mathbf{y}_{t-1} \right]$ are fed into the encoder-decoder module. In this module, the encoder processes the visual features $\mathbf{X}$, which are then fused with a multi-dimensional preference vector $\mathbf{p}$ through the PVF network. These fused features are then combined with the text embeddings and input into the decoder to generate the next token. The MOO module calculates rewards based on multiple evaluation metrics and weights these rewards using preference vectors to formulate a multi-objective reward function. An RL algorithm is then applied to optimize this function, thereby generating reports that align with the specified preferences. Next, we will detail our proposed PVF and MOO modules together with a two-stage training procedure.

\subsection{Encoder-Decoder with the PVF Network} 
The visual features and text embedding are utilized as inputs of the encoder and decoder, respectively. First, the visual features $\mathbf{X}$ are fed into the encoder by Eq. (\ref{eq_encoder}) to obtain the encoded visual features $\mathbf{E}$,
\begin{equation}\label{eq_encoder}
    \mathbf{E}= \left[\mathbf{e}_1,\mathbf{e}_2...,\mathbf{e}_S\right]=f_e(\mathbf{X}),
\end{equation} 
where $f_e$ is the Encoder Layers in the standard Transformer. Then the encoded visual features $\mathbf{E}$ are fed into the PVF network to fuse with the preference vector $\mathbf{p}$.

The key challenge of our MPO approach is how to align the RRG model with multiple preferences. To address this challenge, we formalize the preferences as multi-dimensional vectors and use them as a condition for the RRG model. Now the question is how to incorporate the preference vector into the report generation model. Inspired by the large language model, the radiology image can be viewed as the prompt of the report generation model, and the preference vector is the conditional input. Therefore, we design a preference vector fusion (PVF) network to integrate the preference vector with encoded visual features.

Specifically, given an $m$-dimensional preference vector $\mathbf{p} = (p_1, p_2, \ldots, p_m)$, where $ \mathbf{p} \in {\Delta}_{m-1}$ represents the preference simplex, $p_i\geq 0$ represents the weight for preference dimension $i$, and ${\sum}_{i=1}^m{p_i} = 1$. During training, a preference vector is randomly sampled from the preference simplex and undergoes linear mapping and dimension expansion to ensure the processed preference vector $\mathbf{P}$ has the same dimension as the encoded visual features $\mathbf{E}$, where $\mathbf{P}=Expand(Linear(\mathbf{p}))$. The core of the PVF network is a fusion module based on a multi-head attention mechanism and a residual connection, where the expanded preference vector serves as the query, while the encoded visual features serve as both the key and value. For each head $j$, the fusion process follows the same procedure described as follows:
\begin{equation}
    \mathbf{H}_j=\mathrm{Softmax}\left(\frac{\mathbf{P}\cdot \mathbf{E}^\top}{\sqrt{d}}\right)\cdot\mathbf{E}.
\end{equation}
Then, the attention feature $\mathbf{H}$ is obtained by concatenating $\mathbf{H}_j$ from all heads (the number of heads is 8 in our method). Finally, the attention feature is fused with the encoded visual feature through residual connection to obtain the fused feature $\mathbf{U}$, which is expressed as follows:
\begin{equation}\label{eq_fusion}
    \mathbf{U} = \mathbf{E} + \alpha \mathbf{H},
\end{equation}
where $\alpha$ is a scaling factor. Then the fused features along with the text embedding $\left[ \mathbf{y}_1, \mathbf{y}_2, \ldots, \mathbf{y}_{t-1} \right]$ from previous steps, serve as the input of the Decoder Layers to generate the current output $y_t$ by Eq. (\ref{eq_gen}), 
\begin{equation}\label{eq_gen}
    y_t = f_d(\mathbf{U}, \mathbf{y}_1, \mathbf{y}_2, \ldots, \mathbf{y}_{t-1}),
\end{equation}
where $f_d$ is the Decoder Layers. Given the entire predicted report sequence $\left[y_1, y_2, \ldots, y_T\right] $ and the associated ground truth report $\left[w_1, w_2, \ldots, w_T\right]$, the basic generation loss is:
\begin{equation}\label{loss_g}
    L_{g}  = - \sum_{t=1}^{T} w_t \cdot \log (y_t). 
\end{equation}

\subsection{Multi-objective Optimization via RL} 

The PVF network merely integrates the preference vector into the report generation network, which is not yet aligned with the preferences and therefore cannot control the behavior of the model. To align with multiple preferences, we propose the Multi-Objective Optimization (MOO) module, which includes a multi-dimensional reward function optimized by multi-objective RL. The MOO module uses the preference vector to represent the weight of each preference and linearly combines the multi-dimensional reward with the preference vector. Subsequently, a policy gradient is employed to optimize this weighted reward function, thereby guiding the model to better align with different preferences.

We conceptualize the report generation model as an \textbf{agent} interacting with external \textbf{environment}, which includes visual and textual features, along with a preference vector $\mathbf{p}$. All model parameters, denoted by $\theta$, define a \textbf{policy} ${\pi}_{\theta}$ that guides the \textbf{action}—predicting the next token, i.e., $p(y_t \mid Y_{1:t}, I, \mathbf{p})$. For each report, once the end-of-sequence token is generated, the agent receives several \textbf{rewards} $r$ based on improvements in multiple evaluation metrics. Some of these evaluation metrics focus on the accuracy of text description, such as BLEU\{1-4\}, ROUGE-L, and some focus on clinical efficacy, such as F1 and Recall of disease labels. The number of rewards corresponds to the dimensions of the preference vector. For the $i^{th}$ evaluation metric, the reward for the action at step $t$ is determined by the improvement in generating the next token $y_t$ on that evaluation metric, which is formulated as follows:
\begin{equation}
    r_{t}^{i} = r^i(Y_{1:t}) - r^i(Y_{1:t-1}),
\end{equation}
where $i \in \{1,2,...,m\}$, $Y_{1:t} = \left[y_1, y_2, \ldots, y_t\right]$ and $Y_{1:t-1} = \left[y_1, y_2, \ldots, y_{t-1}\right]$. Therefore, at step $t$,  the multi-objective reward function weighted by the preference vector is calculated as follows:
\begin{equation}
    R_t = \sum_{i=1}^m{p_i r_{t}^{i}} = \sum_{i=1}^m p_i \Big( r^i(Y_{1:t}) - r^i(Y_{1:t-1}) \Big).
\end{equation}

Therefore, the total reward $R$ of generating a report $Y = \left[y_1, y_2, \ldots, y_T\right]$ is the sum of $R_{t}$:
\begin{equation}
\begin{aligned}
     R &= \sum_{t=1}^T{ R_t} =\sum_{t=1}^T \sum_{i=1}^m {p_i \Big( r^i(Y_{1:t}) - r^i(Y_{1:t-1}) \Big)} \\
     & = \sum_{i=1}^m \sum_{t=1}^T{p_i \Big( r^i(Y_{1:t}) - r^i(Y_{1:t-1}) \Big)} = \sum_{i=1}^m {p_i} r^i(Y) ,
\end{aligned}
\end{equation}
where $R$ represents the multi-objective rewards obtained from the sampling strategy. Then the model is trained to maximize the expected reward $\mathbb{E}_{Y \sim \pi_{\theta}} \left[ \sum_{i=1}^m {p_i} r^i(Y)\right]$ from the generated report $Y$ via a sampling strategy. The loss function is now defined as follows:
\begin{equation}
    \mathcal{L}(\theta) = -\mathbb{E}_{Y \sim \pi_{\theta}} \left[ \sum_{i=1}^m {p_i} r^i(Y) \right].
\end{equation}
The gradient of the loss function $\mathcal{L}(\theta)$ is computed using the REINFORCE algorithm \cite{williams1992simple} as follows:
\begin{equation}
    \nabla_{\theta} \mathcal{L}(\theta) = -\mathbb{E}_{Y \sim {\pi}_{\theta}} \left[ \Big(\sum_{i=1}^m {p_i} r^i(Y)\Big) \nabla_{\theta} \log {\pi}_{\theta}(Y) \right].
    \label{eq:sample_grad}
\end{equation}
To stabilize the RL training process, we follow \citet{rennie2017self} and introduce a reference reward $b$, where $b$ is a weighted sum of multiple metrics calculated via greedy sampling. Thus, Eq. (\ref{eq:sample_grad}) can be re-formulated as follows:
\begin{equation}
    \nabla_{\theta} \mathcal{L}(\theta) = -\mathbb{E}_{Y \sim {\pi}_{\theta}} \left[ \Big( \sum_{i=1}^m {p_i} r^i(Y) - b \Big) \nabla_{\theta} \log {\pi}_{\theta}(Y) \right].
\end{equation}
Then, we approximate the expected gradient through Monte-Carlo sampling, which averages the gradient of $N$ generated reports $\{Y_i\}$ sampled from $\pi_\theta$ as follows:
\begin{equation}\label{eq_update}
     \nabla_{\theta} \mathcal{L}(\theta) \approx - \frac{1}{N} \sum_{i=1}^N \left[ \Big( \sum_{i=1}^m {p_i} r^i(Y_i) - b \Big) \nabla_{\theta} \log {\pi}_{\theta}(Y_i)\right].
\end{equation}

\subsection{Training}
MPO comprises two training stages: the first employs the primary generation loss in Eq. (\ref{loss_g}) to regularize the action space, and the second involves multi-objective optimization based on Eq. (\ref{eq_update}). During training, we randomly sample diverse preference vectors from the preference space and refine the RRG model by optimizing weighted multi-objective rewards. This strategy achieves an optimal policy across the entire preference space, ensuring that the optimal model adapts to various preference conditions.  During inference, given a preference vector, our model can generate reports that cater to this preference without further fine-tuning.
\section{Experiments}
\subsection{Experiment Settings}
\textbf{Datasets. } We evaluate our method on two public datasets, IU-Xray \cite{demner2016preparing} and MIMIC-CXR \cite{johnson2019mimic}. IU-Xray consists of 7,470 chest X-ray images, accompanied by 3,955 reports. Following CMN \cite{R2GenCMN}, only the findings and impressions are included. Datasets are randomly split into 7:1:2 for train, val, and test. MIMIC-CXR, the largest publicly available dataset for RRG, contains 337,110 chest X-ray images and 227,835 corresponding reports. We adhere to the official dataset splits to ensure a fair comparison.

\textbf{Evaluation Metrics. } 
We utilize the widely used Natural Language Generation (NLG) metrics including BLEU\{1-4\} \cite{DBLP:conf/acl/PapineniRWZ02}, METOR, and ROUGE-L \cite{lin2004rouge} to assess the quality of the generated text reports. Additionally, to evaluate Clinical Efficacy (CE), we use CheXbert \cite{CheXbert} to annotate labels for 14 observations in medical reports. Precision (P), recall (R), and F1 scores are then calculated based on the labels according to the label between the reference and generated reports. Higher values in both NLG and CE metrics indicate better performance.

\textbf{Implementation Details. } 
Our method was implemented in PyTorch and trained on an NVIDIA 4090 GPU with 24GB of memory. We employ ResNet101, pre-trained on ImageNet, as the visual extractor, and a randomly initialized Transformer \cite{vaswani2017attention}, as the encoder-decoder. The initial learning rate for ResNet101 and remaining networks are $1 \times 10^{-6}$ and $1 \times 10^{-5}$, respectively. We use the Adam optimizer for training and include a beam search of width 3. Maximum report lengths are set to 60 words for IU-Xray and 100 words for MIMIC-CXR. Our model undergoes Maximum Likelihood Estimation(MLE) training for 50 epochs on IU-Xray and 30 epochs on MIMIC-CXR to regularize the action space, followed by an RL training phase using the same optimizer. During training, the sampling interval of the preference vector on both datasets is 0.1, For the IU-Xray dataset, we use NLG metrics as the multi-objective rewards, the batch size is 8, and $\alpha=3$. For the MIMIC-CXR dataset, we combine NLG and CE metrics as the multi-objective rewards, the batch size is 6, and $\alpha=0.5$.

\subsection{Effective of Preference Guidance}
To verify the effectiveness of preference guidance, we test the model trained in a two-dimensional preference vector space. (The results of the three-dimensional preference space are detailed in the \textcolor{blue}{supplementary material}.) Tables \ref{tab_preference_iu} and \ref{tab_preference_cxr} show the test results with a preference vector sampling interval of 0.2 on the IU-Xray and MIMIC-CXR datasets, respectively. For the IU-Xray dataset, the two rewards in the multi-objective optimization are calculated from the BLEU-1 and ROUGE-L metrics, where $p_1$ and $p_2$ represent their weights, respectively. For the MIMIC-CXR dataset, the two rewards are calculated from BLEU-1 in the NLG metric and F1 in the CE metric. From Table \ref{tab_preference_iu}, as the weight of BLEU-1 decreases from 1 to 0, the weight of ROUGE-L increases from 0 to 1; the B1 score decreases from 0.548 to 0.531 (the other three BLEU scores also gradually decrease), while the ROUGE-L score increases from 0.413 to 0.415. In Table \ref{tab_preference_cxr}, with this preference configuration, the B1 score decreases from 0.416 to 0.392, and other NLG metrics also gradually decrease, while the F1 score significantly improves from 0.363 to 0.436. Since F1 is calculated based on precision and recall, the change trends of P and R are consistent with F1 and also gradually increase. The above observations demonstrate that our MPO method can customize the behavior of the model by controlling the preference vector, thereby generating reports that align with specific preferences. 

\begin{table}[H]
\begin{center}\renewcommand{\arraystretch}{1.1} 
\begin{tabular}{cc|cccc|c}\hline
$p_1$   & $p_2$   & B1   & B2   & B3    & B4    & RG-L   \\ \hline
1   & 0   &\textbf{0.548} & \textbf{0.383} & \textbf{0.278} & \textbf{0.209} & 0.413  \\
0.8 & 0.2 & 0.546 & 0.381 & 0.276 & 0.207 & 0.4137 \\
0.6 & 0.4 & 0.544 & 0.379 & 0.274 & 0.206 & 0.4139 \\
0.4 & 0.6 & 0.542 & 0.377 & 0.272 & 0.205 & 0.414  \\
0.2 & 0.8 & 0.537 & 0.374 & 0.270  & 0.203 & 0.4143 \\ 
0   & 1   & 0.531 & 0.37  & 0.268 & 0.200   & \textbf{0.415}  \\ \hline
\end{tabular}\caption{Test results on the IU-Xray dataset under different preference vectors, where $p_1$ and $p_2$ represent the weights of the rewards based on B1 and RG-L, respectively.}\label{tab_preference_iu}
\end{center}\end{table}
\vspace{-1mm}

\begin{table*}[htp!]
\begin{center}\renewcommand{\arraystretch}{0.9} 
\begin{tabular}{cc|cccccc|ccc}\hline
$p_1$  & $p_2$   & \small{B1}    & \small{B2}    & \small{B3}    & \small{B4}    & \small{MTR}   & \small{RG-L}  & \small{P}   & \small{R}     & \small{F1}\\ \hline
1   & 0   & \textbf{0.416} & \textbf{0.269} & \textbf{0.191} & \textbf{0.139} & \textbf{0.162} & \textbf{0.309} & 0.363 & 0.315 & 0.316 \\
0.8 & 0.2 & 0.413 & 0.267 & 0.188 & 0.137 & 0.161 & 0.308 & 0.375 & 0.319 & 0.318 \\
0.6 & 0.4 & 0.407 & 0.262 & 0.183 & 0.134 & 0.158 & 0.305 & 0.393 & 0.363 & 0.329 \\
0.4 & 0.6 & 0.406 & 0.26  & 0.182 & 0.132 & 0.158 & 0.305 & 0.401 & 0.363 & 0.330 \\
0.2 & 0.8 & 0.405 & 0.259 & 0.181 & 0.131 & 0.157 & 0.303 & 0.412 & 0.364 & 0.335 \\ 
0   & 1   & 0.392 & 0.249 & 0.173 & 0.125 & 0.151 & 0.299 & \textbf{0.436} & \textbf{0.376} & \textbf{0.353} \\ \hline
\end{tabular}\caption{Test results on the MIMIC-CXR dataset under different preference vectors, where $p_1$ and $p_2$ correspond to the weights of the reward function based on B1 and F1, respectively.} \label{tab_preference_cxr}
\end{center}\end{table*}
\vspace{3mm}

\begin{table*}[!ht]
    \begin{center}\renewcommand{\arraystretch}{1.1} 
    \resizebox{\textwidth}{!}{
    \begin{tabular}{c|c|c|cccccc|cccccc}\hline
        \multirow{2}*{Methods} & \multirow{2}*{Years} & \multirow{2}*{Meeting} & \multicolumn{6}{c}{\textbf{IU-Xray}} & \multicolumn{6}{c}{\textbf{MIMIC-CXR}}\\ \cline{4-15} 
        ~ & ~ & ~ & B1 & B2 & B3 & B4 & MTR & RG-L & B1 & B2 & B3 & B4 & MTR & RG-L \\ \hline
        R2Gen & 2020 & EMNLP & 0.470 & 0.304 & 0.219 & 0.165 & 0.187 & 0.371 & 0.353 & 0.218 & 0.145 & 0.103 & 0.128 & 0.267 \\ 
        CMN   & 2021 & ACL & 0.475 & 0.309 & 0.222 & 0.170 & 0.191 & 0.375 & 0.353 & 0.218 & 0.148 & 0.108 & 0.142 & 0.277 \\ 
        CA    & 2021 & ACL & 0.492 & 0.314 & 0.222 & 0.169 & 0.193 & 0.381 & 0.350 & 0.219 & 0.152 & 0.109 & 0.151 & 0.283 \\
        CMN+RL & 2022 & ACL &0.494 & 0.321 & 0.235 & 0.181 & 0.201 & 0.384 & 0.381 & 0.232 & 0.155 & 0.109 & 0.151 & 0.287 \\ 
        XPRONET & 2022 & ECCV & 0.525 & 0.357 & 0.262 & 0.199 & 0.220 & 0.411 & 0.344 & 0.215 & 0.146 & 0.105 & 0.138 & 0.279 \\ 
        KiUT & 2023 & CVPR & 0.525 & 0.360 & 0.251 & 0.185 & \textbf{0.242} & 0.409 & 0.393 & 0.243 & 0.159 & 0.113 & 0.160 & 0.285 \\ 
        METrans & 2023 & CVPR & 0.483 & 0.322 & 0.228 & 0.172 & 0.192 & 0.380 & 0.386 & 0.250 & 0.169 & 0.124 & 0.152 & 0.291 \\ 
        MMTN & 2023 & AAAI &0.486 & 0.321 & 0.232 & 0.175 & - & 0.375 & 0.379 & 0.238 & 0.159 & 0.116 & 0.161 & 0.283 \\ 
        MAN & 2024 & AAAI &0.501 & 0.328 & 0.230 & 0.170 & 0.213 & 0.386 & 0.396 & 0.244 & 0.162 & 0.115 & 0.151 & 0.274 \\ 
        COMG+RL & 2024 & WACV &0.536 & 0.378 & 0.275 & 0.206 & 0.218 & 0.383 & 0.363 & 0.235 & 0.167 & 0.124 & 0.128 & 0.290 \\ 
        MPO & Ours & AAAI & \textbf{0.548} & \textbf{0.383} & \textbf{0.278} & \textbf{0.209} & \underline{0.224} & \textbf{0.415} & \textbf{0.416} & \textbf{0.269} & \textbf{0.191} & \textbf{0.139} & \textbf{0.162} & \textbf{0.309} \\ \hline
    \end{tabular}
    }
    \caption{Comparison with previous works on IU-Xray and MIMIC-CXR using NLG metrics. The best values are in bold.}\label{tab_previous_nlg}
    \end{center}
\end{table*}

\begin{figure}[ht]
\centering \includegraphics[width=0.45\textwidth]{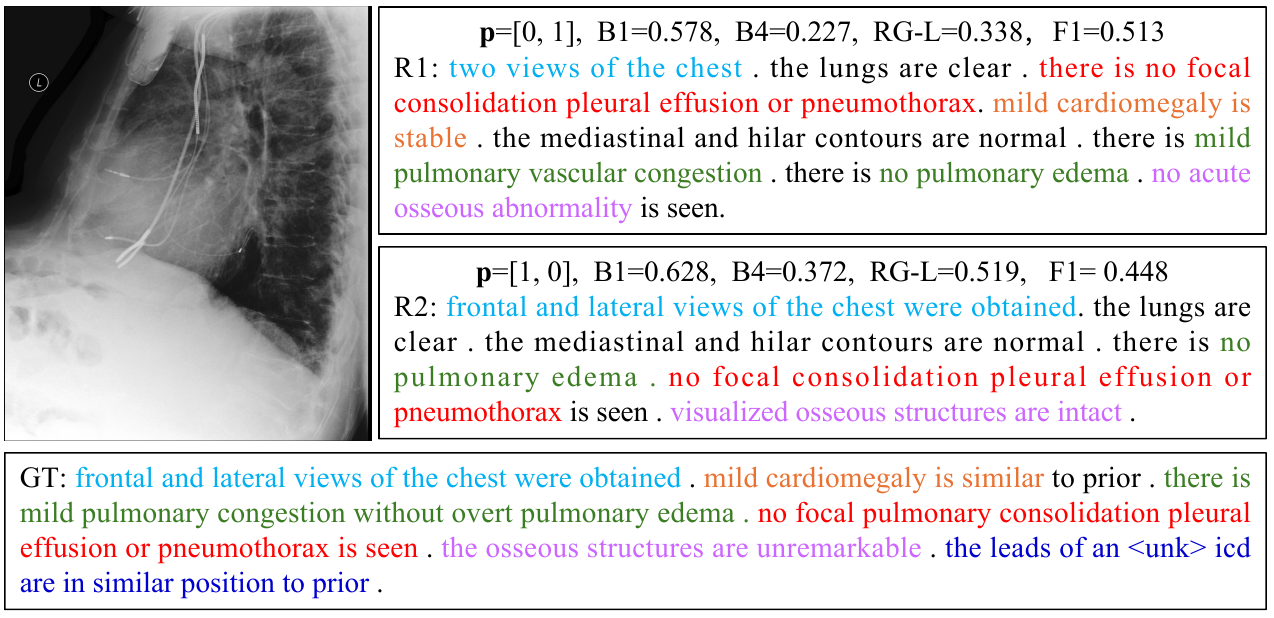}
\caption{Reports from ground truth and MPO with different preference configurations on MIMIC-CXR, where the same color highlights the descriptions of the same content.}
\label{fig2}
\end{figure}

\noindent\textbf{ Qualitative results. }
To further verify the effectiveness of preference guidance intuitively, we perform qualitative analysis on the MIMIC-CXR dataset. Figure \ref{fig2} shows the comparison between the reference report ``GT'' and reports generated under two preference configurations. ``R1'' is the generated report when the preference vector is [0, 1], indicating a complete preference for CE metrics. ``R2'' is the generated report when the preference vector is [1, 0], indicating a complete preference for NLG metrics. 
Both R1 and R2 mention the absence of focal pulmonary consolidation, pleural effusion, pneumothorax, and pulmonary edema, and describe the imaging views as ``frontal and lateral views of the chest'', consistent with the GT. For the difference, GT mentions ``mild pulmonary congestion without overt pulmonary edema'' reflected in R1 as ``mild pulmonary vascular congestion'' while R2 lacks this detail. Both GT and R1 recognize ``mild cardiomegaly'' as stable, but missing in R2. The above observations indicate that R1 aligns more closely with GT in terms of content consistency and medical terminology. Conversely, R2 has a high degree of overlap in text descriptions but omits critical details like cardiomegaly, which are essential in GT. These observations are consistent with the preference vector, indicating that the preference vector can effectively control the performance of report generation.

\begin{table}[ht]
\begin{center}\renewcommand{\arraystretch}{1.1} 
\begin{tabular}{c|c|ccc}\hline
        Methods & Years & P & R & F1  \\ \hline
        R2Gen & 2020 & 0.333 & 0.273 & 0.276  \\ 
        CMN & 2021 & 0.334 & 0.275 & 0.278  \\ 
        CA & 2021 & 0.352 & 0.298 & 0.303  \\ 
        CMN+RL & 2022 & 0.342 & 0.294 & 0.292  \\ 
        KiUT & 2023 & 0.371 & 0.318 & 0.321  \\ 
        METrans & 2023 & 0.364 & 0.309 & 0.311  \\
        MAN & 2024 & 0.411 & \textbf{0.398} & \textbf{0.389}  \\
        COMG+RL & 2024 & 0.424 & 0.291 & 0.345  \\
        MPO & ours & \textbf{0.436} & \underline{0.376} & \underline{0.353} \\ \hline
\end{tabular}\end{center}
\caption{Comparison of clinical metrics and diversity scores on MIMIC-CXR. The best results are highlighted in bold.}\label{tab_ce}
\end{table}
\vspace{0mm}

\begin{table*}[!ht]
    \begin{center}\renewcommand{\arraystretch}{1.2} 
    \resizebox{1.0\linewidth}{!}{\begin{tabular}{c|cccccc|ccccccccc}\hline
\multirow{2}*{Methods }& \multicolumn{6}{c}{\textbf{IU-Xray}} & \multicolumn{9}{c}{\textbf{MIMIC-CXR}} \\ \cline{2-16}
~  &B1   & B2 & B3 & B4 & MTR & RG-L & B1 & B2 & B3 & B4 & MTR & RG-L & P & R & F1 \\  \hline
Base   & 0.470  & 0.304   & 0.219    & 0.165  & 0.187   & 0.371  &0.353 & 0.218 & 0.145   & 0.103   & 0.128    & 0.267    & 0.333    & 0.273    & 0.276 \\
+MOO   & 0.540  & 0.376   & 0.273    & 0.202  & 0.220   & 0.408  &0.381 & 0.243 & 0.168   & 0.122  & 0.144     & 0.298    & 0.372    & 0.358    & 0.312 \\
+MOO+PVF & \textbf{0.548} & \textbf{0.383} & \textbf{0.278} & \textbf{0.209} & \textbf{0.224} & \textbf{0.415} & \textbf{0.416} & \textbf{0.269} & \textbf{0.191} & \textbf{0.139} & \textbf{0.162} & \textbf{0.309} & \textbf{0.436} & \textbf{0.376} & \textbf{0.353} \\ \hline
    \end{tabular}}
    \caption{Results of ablation experiments on IU-Xray and MIMIC-CXR datasets. The best results are highlighted in bold.}\label{tab_ablation}
    \end{center}
\end{table*}
\vspace{-5mm}

\subsection{Comparison with Previous Works}
We compare MPO with a wide range of SOTA methods in the past five years, including R2Gen \cite{R2Gen}, CMN \cite{R2GenCMN}, CA \cite{CA}, CMN+RL \cite{qin2022reinforced}, XPRONET \cite{XPRONET}, KiUT\cite{KIUT}, METrans \cite{Metransfor}, MMTN \cite{MMTN}, MAN \cite{MAN} and COMG+RL \cite{COMG}. The results of other methods are cited from their papers. The NLG results on both datasets are shown in Table \ref{tab_previous_nlg}, and the clinical efficacy results on the MIMIC-CXR are shown in Table \ref{tab_ce}. 

Overall, our proposed MPO achieves the best results on all NLG metrics on both datasets, except for the MTR metric in the IU-Xray dataset, which demonstrates that the proposed MPO method can not only generate reports that cater to human preferences but also generate higher-quality reports. Comparing the CE metrics in Table \ref{tab_ce}, our method achieves the best performance in precision and the second-best performance in recall and F1 scores. Although MAN has higher recall and F1 scores, it cannot adapt to the preferences of radiologists. Additionally, our MPO significantly outperforms the baseline method R2Gen \cite{R2Gen} in all metrics. Specifically, improving Precision from 0.333 to 0.436, Recall from 0.273 to 0.376, and F1 from 0.276 to 0.353. The average value of CE is improved by about 9.4\%, demonstrating our approach's effectiveness. 

\subsection{Other Results and Analyses}
\subsubsection{Ablation Study} 
Ablation studies are performed on both IU-Xray and MIMIC-CXR datasets to explore the impact of each component in our MPO method. Three variants were investigated: (1) ``Base'' represents our baseline, R2Gen \cite{R2Gen}. (2) ``Base+MOO'' combines the baseline with our MOO module. It uses a preference vector to weigh multiple rewards and optimizes it via multi-objective RL. (3) ``Base+MOO+PVF'' represents our full model, which includes both the PVF and MOO modules. The results are shown in Table \ref{tab_ablation}. “Base+MOO” achieves significant improvements across all metrics on both datasets compared to ``Base'', demonstrating that optimizing multiple objective reward functions can significantly improve model performance. Furthermore, ``Base+MOO+PVF'' outperforms ``Base+MOO'' in all metrics, suggesting that incorporating preference vectors as conditional inputs and utilizing multi-objective optimization can further improve the performance of report generation.

\begin{table}[H]
\begin{center}\setlength{\tabcolsep}{3.5pt}
\renewcommand{\arraystretch}{1.1} 
\begin{tabular}{ccccccc}\hline
Fusion &B1        &B2        &B3       &B4       &MTR       &RG-L  \\ \hline
Concat &0.544    &0.379    &0.275     &0.206    &0.222     &0.411  \\
Add    &0.541    &0.378   &0.275      &0.208    &0.222     &0.412    \\
Mul    &0.543    &0.378   &0.274      &0.205    &0.222     &0.410    \\
PVF  &\textbf{0.548} & \textbf{0.383} & \textbf{0.278} & \textbf{0.209} & \textbf{0.224} & \textbf{0.415} \\ \hline
\end{tabular}\caption{Results of different fusion methods on IU-Xray.}\label{tab_f_methods}
\end{center}\end{table}
\vspace{-4mm}

\subsubsection{Fusion Methods} \label{Fusion methods}
In addition to PVF, we explore three other fusion methods to fuse preference vectors. The first concatenates (Concat) the expanded preference vector with the encoded visual features and then applies a linear transformation to obtain the fused features. The second and third methods align the dimension of the preference vector with the encoded visual features through linear transformation and dimensionality expansion, followed by either addition (Add) or dot multiplication (Mul), respectively. As shown in Table \ref{tab_f_methods}, our PVF network outperforms other methods across all evaluation metrics. The Concat, Add, and Mul methods perform poorly because they may not capture the complex relationships between preference vectors and encoded visual features through simple fusion. In contrast, the PVF network leverages the attention mechanism to align the preference vector with key information in the visual features, resulting in more effective fusion and superior performance.
\begin{table}[ht]
\renewcommand{\arraystretch}{1.0}
\begin{tabular}{ccccccc}\hline
$\alpha$    & \small{B1}    & \small{B2}      & \small{B3}        & \small{B4}             & \small{MTR}            & \small{RG-L}      \\ \hline
0.1    & 0.539    & 0.375   & 0.273      & 0.206          & 0.221          & 0.409     \\
0.5    & 0.54     & 0.375   & 0.271      & 0.203          & 0.219          & 0.409     \\
1      & 0.541    & 0.376   & 0.272      & 0.204          & 0.221          & 0.408    \\
3    & \textbf{0.548} & \textbf{0.383} & \textbf{0.278} & \textbf{0.209} & \textbf{0.224} & \textbf{0.415} \\
5     & 0.544     & 0.38    & 0.277      & 0.207          & 0.222          & 0.413    \\
10    & 0.546     & 0.381   & 0.277      & 0.208          & 0.222          & 0.411  \\ \hline
\end{tabular}\caption{Hyperparameter analysis of $\alpha$ on IU-Xray.}\label{tab_hyper}
\end{table}
\vspace{-3mm}

\subsubsection{Hyperparameter Analysis}
Hyperparameter analysis of $\alpha$ is conducted on the IU-Xray dataset. The candidate set for $\alpha$ includes \{0.1, 0.5, 1, 3, 5, 10\}. The experimental results in Table \ref{tab_hyper}, indicate that the value of $\alpha$ within a reasonable range does not significantly affect the results. The best performance is achieved when $\alpha=3$.

\section{Conclusion}
In this study, we propose a novel method for automatic RRG via Multi-objective Preference Optimization (MPO), which effectively addresses the inherent heterogeneity and multi-dimensionality of radiologists’ preferences. We propose two innovative modules: the preference vector fusion network and the multi-objective optimization module to enable conditional generation and effective preference alignment. 
By employing preference vectors to condition the RRG model and optimizing the preference-weighted multi-dimensional reward function through reinforcement learning, our model derives an optimal policy over the entire preference space, resulting in a model with high adaptability to different preferences. Through extensive experiments on two public datasets, we have demonstrated that our method can cater to various preferences within a single model and achieve state-of-the-art performance. In future work, we will explore designing new evaluation metrics as reward functions for preference alignment.

\section{Acknowledgement}

This work is supported by the National Natural Science Foundation of China under Grant No. 62306115 and No. 62476087.

\bibliography{aaai25}

\end{document}


\maketitle

\section{Effective of Preference Guidance}
To further verify the effectiveness of preference-guided report generation, we test the model trained in a three-dimensional preference vector space. Tables \ref{tab_preference_iu} and \ref{tab_preference_cxr} show the test results under different preference vector configurations on the IU-Xray and MIMIC-CXR datasets, respectively. For the IU-Xray dataset, the three rewards in the multi-objective optimization are calculated from the BLEU-1, BLEU-4, and ROUGE-L, with $p_1$, $p_2$, and $p_3$ representing their respective weights. For the MIMIC-CXR dataset, the three rewards are calculated from BLEU-1 and BLEU-4 in the NLG metrics, and F1 in the CE metric. Each table shows four groups of preference configurations: the first three rows represent the results of fully preferring one of the metrics, while the last group assigns equal weights to all metrics.

 \begin{table}[h]
\begin{center}\renewcommand{\arraystretch}{1.0} 
\begin{tabular}{ccc|cccc|c}\hline
$p_1$ & $p_2$ &$p_3$   & B1     & B4    & RG-L   \\ \hline
1 & 0 & 0 &\textbf{0.548} & 0.207 & 0.412  \\
0 & 1 & 0 & 0.545 & \textbf{0.209} & 0.412\\
0 & 0 & 1 & 0.532 & 0.201 & \textbf{0.414} \\
1/3 & 1/3 & 1/3 & \underline{0.546} & \underline{0.208} & \underline{0.413}  \\ \hline
\end{tabular}\caption{Test results on IU-Xray under 3-dimensional preference vectors, where $p_1$, $p_2$ and and $p_3$ correspond to the weights of the reward function based on B1, B4 and RG-L, respectively.}\label{tab_preference_iu}
\end{center}\end{table}

\begin{table}[h]
\begin{center}\renewcommand{\arraystretch}{1.0} 
\begin{tabular}{ccc|cc|ccc}\hline
$p_1$ & $p_2$ &$p_3$   & B1      & B4     & P   & R     & F1\\ \hline
1 & 0 & 0 & \textbf{0.404} & 0.133 & 0.382 & 0.395 & 0.386 \\
0 & 1 & 0 & 0.401  & \textbf{0.136} & 0.386 & 0.396 & 0.385 \\
0 & 0 & 1 & 0.374  & 0.121 & \textbf{0.456} & \textbf{0.505} & \textbf{0.465} \\
1/3 & 1/3 & 1/3 & \underline{0.402}  & \underline{0.135} & \underline{0.388} & \underline{0.397} & \underline{0.388} \\\hline
\end{tabular}\caption{Test results on MIMIC-CXR under different preference vectors, where $p_1$, $p_2$ and $p_3$ correspond to the weights of the reward function based on B1, B4 and F1, respectively.} \label{tab_preference_cxr}
\end{center}\end{table}

From Table \ref{tab_preference_iu}, we can see that when the preference weight of a certain metric is set to 1, the corresponding metric achieves the best performance. This demonstrates that the preference vector can effectively guide the model’s behavior. Similar observations can be made on the MIMIC-CXR dataset, where the trends of Precision (P) and Recall (R) are consistent with the F1 score, as F1 is calculated based on precision and recall. Additionally, when the preference weights for each metric are equal, it indicates that each evaluation metric is equally important. In this scenario, the model achieves the second-best result across all metrics, resulting in an average model with equal preference.

\begin{table}[h]
\begin{center}\renewcommand{\arraystretch}{1.0} 
\setlength{\tabcolsep}{0.14cm} 
\begin{tabular}{ccc|ccccccc|c}\hline
$p_1$ & $p_2$ &$p_3$   & B1     & B4    & P   & R & F & CliQ  \\ \hline
1 & 0 & 0 &\textbf{0.423} & \textbf{0.139} & 0.384  & 0.382 & 0.381 & 2.63 \\
0 & 1 & 0 & 0.372 & 0.125 & \textbf{0.448} & \textbf{0.474} & \textbf{0.442} & 2.66\\
0 & 0 & 1 & 0.392 & 0.128 & 0.393 & 0.391 & 0.387 & \textbf{2.81} \\
1/3 & 1/3 & 1/3 & \underline{0.403} & \underline{0.134} & \underline{0.443} & \underline{0.456} & \underline{0.432} & \underline{2.68}  \\ \hline
\end{tabular}\caption{Test results on MIMIC-CXR under different preference vectors, where $p_1$, $p_2$ and $p_3$ correspond to the weights of the reward function based on B1, F1 and RadCliQ, respectively.}\label{tab_preference_cliq}
\end{center}\end{table}

We included RadCliQ as an additional CE metric to enhance the credibility further. The results in Table\ref{tab_preference_cliq} show that setting a metric’s preference weight to 1 yields the best performance for that metric. When preference weights for each metric are equal, it indicates equal importance for all evaluation metrics. In this scenario, the model achieves the second-best across all metrics, resulting in an average model with balanced preferences. This shows that the preference vector can effectively guide the model’s behavior.

\begin{figure}[ht]
\centering \includegraphics[width=0.45\textwidth]{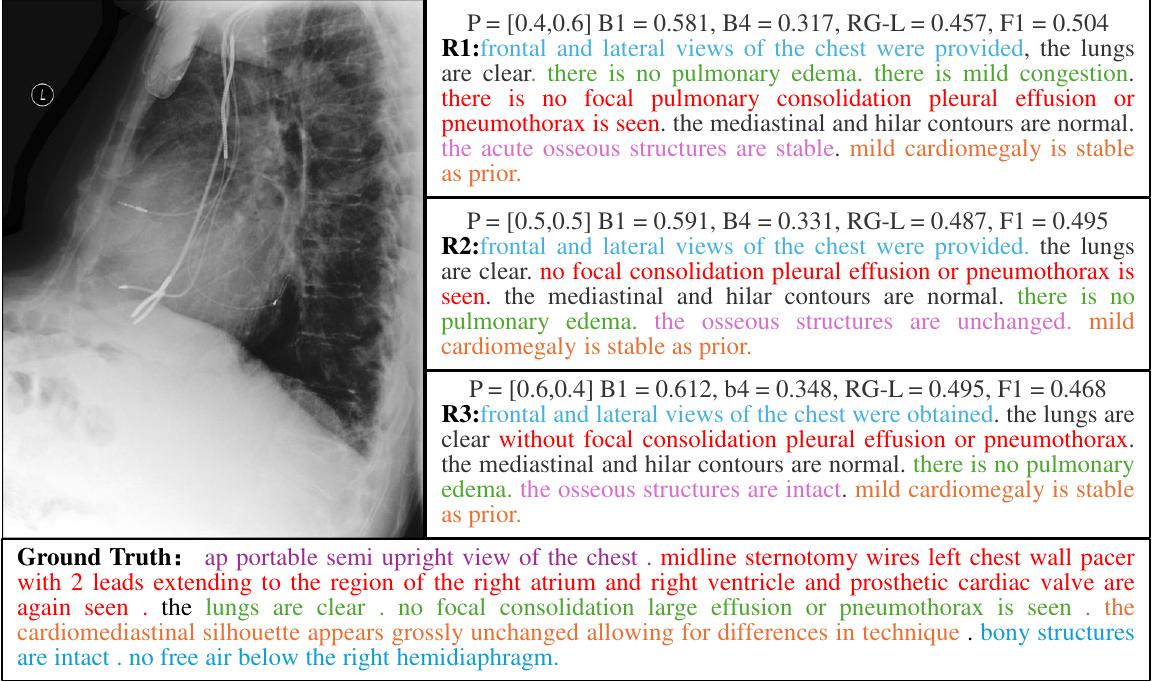}
\caption{Reports from ground truth and MPO with different preference configurations on MIMIC-CXR, where the same color highlights the descriptions of the same content.}
\label{fig3}
\end{figure}
\vspace{0.7cm}

We add additional reports generated with $p = \{[0.5,0.5], [0.4,0.6], [0.6,0.4]\}$, as shown in Figure \ref{fig3}.The evaluation metrics show a consistent trend with their corresponding preference values. From the NLG view, all generated reports maintain grammatically correct structures, but R3 closely mimics the GT structure by using terminology like "were obtained".From the CE view, R1 identifies "mild cardiomegaly" and "mild congestion", aligning well with GT’s mentions, while R2 and R3 do not mention pulmonary "congestion". Thus, R1 shows better CE, while R3 has better NLG.

\begin{table*}[ht]
\begin{center}\renewcommand{\arraystretch}{1.1} 
\setlength{\tabcolsep}{0.14cm} 
\begin{tabular}{cccccccc}
\hline
Model  & \#Params(M) & FLOPs(G) & \begin{tabular}[c]{@{}c@{}}IU-Xray \\ GPU(GB)\end{tabular} & \begin{tabular}[c]{@{}c@{}}IU-Xray training \\ time(s/epoch)\end{tabular} & \begin{tabular}[c]{@{}c@{}}IU-Xray inference\\  time(ms)\end{tabular} & \begin{tabular}[c]{@{}c@{}}MIMIC training \\ time(h/epoch)\end{tabular} & \begin{tabular}[c]{@{}c@{}}MIMIC inference \\ time(ms)\end{tabular} \\ \hline
R2Gen  & 78.5        & 224.6    & 4.46                                                       & 55                                                                        & 72.45                                                                 & 3.51                                                                    & 163.13                                                              \\ \hline
CMN    & 59.1        & 218.84   & 8.56                                                       & 40                                                                        & 37.18                                                                 & 6.41                                                                    & 63.37                                                               \\ \hline
CMN+RL & 60.75       & 219.97   & 19.58                                                      & 249                                                                       & 36.64                                                                 & 19.63                                                                   & 63.35                                                               \\ \hline
Ours   & 63.26       & 221.74   & 20.58                                                      & 326                                                                       & 48.02                                                                 & 19.18                                                                   & 79.21                                                               \\ \hline

\end{tabular}
\caption{Comparison of the number of parameters, FLOPs, GPU memory usage, and training/inference time on the IU-Xray and MIMIC-CXR datasets between our model and other comparative methods.}
\label{table_trainging}
\end{center}
\end{table*}

\section{Complexity and Efficiency Analysis}
The Table \ref{table_trainging} shows the computational cost required for methods with public code, with batch sizes of 8 and 6 for IU-Xray and MIMIC-CXR datasets, respectively. The two-stage methods (CMN+RL, Ours) require more GPUs and computation time than one-stage methods (R2Gen, CMN). Unlike the recent two-stage model CMN+RL, our method requires slightly more training time because we learn a multi-objective policy that can generate a set of optimal solutions in a continuous multi-objective preference space, while previous works optimize a single reward that is tailored to a specific weighted preference.